\documentclass{article}

\PassOptionsToPackage{numbers, compress}{natbib}

\usepackage[preprint]{neurips_2025}
\usepackage{graphicx}
\usepackage{float} 
\usepackage{subcaption} 
\usepackage{wrapfig} 
\usepackage{xcolor}
\usepackage{soul} 
\setul{0.5ex}{0.25ex}

\usepackage[utf8]{inputenc} 
\usepackage[T1]{fontenc}    
\usepackage{hyperref}       
\usepackage{url}            
\usepackage{booktabs}       
\usepackage{amsfonts}       
\usepackage{nicefrac}       
\usepackage{microtype}      
\usepackage{xcolor}         

\title{FormCoach: Lift Smarter, Not Harder}

\author{
  Xiaoye Zuo \\
  University of Pennsylvania \\
  \texttt{zuoxy@seas.upenn.edu} \\
  \And
  Nikos Athanasiou \\
  Max Planck Institute for Intelligent Systems \\
  \texttt{nathanasiou@tuebingen.mpg.de} \\
  \And
  Ginger Delmas \\
  \texttt{ginger.delmas@naverlabs.com} \\
  \And
  Yiming Huang \\
  University of Pennsylvania \\
  \texttt{ymhuang9@seas.upenn.edu} \\
  \And
  Xingyu Fu \\
  University of Pennsylvania\\
  \texttt{xingyuf2@seas.upenn.edu} \\
  \And
    Lingjie Liu \\
  University of Pennsylvania \\
  \texttt{lingjie.liu@seas.upenn.edu} \\
}

\begin{document}

\maketitle

\begin{abstract}
Good form is the difference between strength and strain, yet for the fast-growing community of at-home fitness enthusiasts, expert feedback is often out of reach. FormCoach transforms a simple camera into an always-on, interactive AI training partner, capable of spotting subtle form errors and delivering tailored corrections in real time, leveraging vision-language models (VLMs). We showcase this capability through a web interface and benchmark state-of-the-art VLMs on a dataset of 1,700 expert-annotated user–reference video pairs spanning 22 strength and mobility exercises. To accelerate research in AI-driven coaching, we release both the dataset and an automated, rubric-based evaluation pipeline, enabling standardized comparison across models. Our benchmarks reveal substantial gaps from human-level coaching, underscoring both the challenges and opportunities in bringing nuanced, context-aware movement analysis into interactive AI systems. By framing form correction as a collaborative, creative process between human and machine, FormCoach opens a new frontier in embodied AI.

\end{abstract}

\section{Introduction}
\label{sec:intro}

The only thing worse than not lifting weights is lifting them with bad form~\cite{Jones_Cowan_Knapik_1994}.  Poor techniques not only diminishes training effectiveness, it can also lead to serious injury~\cite{bjsm2016_sysreview}. 
Yet, most individuals lack continuous access to a personal coach who can spot and correct errors on the spot~\cite{Tsiouris_Tsakanikas_Gatsios_Fotiadis_2020, plosone2024coach_behavior}.  
Without timely, precise feedback,  
at-home fitness enthusiasts risk reinforcing bad habits, limiting progress, and an increasing likelihood of harm~\cite{Graves_Iyer_Willis_Ebel_Rivara_Vavilala_2013}.  
We introduce \emph{FormCoach}, an interactive AI form coach designed to address exactly this challenge.
FormCoach transform a simple camera, whether in a smartphone, tablet, or smart mirror, into an always-available form coach. Users can specify personalized goals such as "focus on my knee alignment," perform an exercise, and receive immediate, concise feedback. By comparing the user's movement in real time against an expert reference video, FormCoach delivers actionable corrections tailored to the individual. 

Previous AI trainer systems~\cite{fieraru2021aifit, Chen_Yang_2020, Liu_Saquib_Zhutian_Kazi_Wei_Fu_Tai_2024, Dittakavi_Bavikadi_Desai_Chakraborty_Reddy_Balasubramanian_Callepalli_Sharma_2022} largely focused on pose estimation and joint angle differences, overlooking richer contextual cues such as body-equipment interaction and ground contact. 
Motion-language models have made significant progress in motion captioning, generation, and editing~\cite{athanasiou2024motionfix, Zhang_Huang_Liu_Tang_Lu_Chen_Bai_Chu_Yu_Ouyang_2024, motionllm}. Yet their progress is constrained by the scarcity of high-quality motion-language datasets, and they rarely address the problem of delivering concise, context-aware comparative feedback. Recent advances in vision-language models (VLMs) have unlocked impressive capabilities in video captioning, question answering, and multimodal reasoning~\cite{chatgpt2025, damonlpsg2025videollama3,chen2024internvl, qwen2}.
However, they have rarely been tested on nuanced, comparative analyses of human motion \cite{Cho_Lin_Srinivasan_Saxon_Kwon_Chavez_May_2025}. 
FormCoach bridges this gap by evaluating
state-of-the-art VLMs on a curated dataset of user-reference exercise pairs, probing their ability to identify and articulate form differences with clarity and precision. In doing so, FormCoach reframes AI as an interactive, creative partner, one that collaborates with users to elevate the at-home
fitness experience.
\begin{figure}
    \centering    \includegraphics[width=0.7\linewidth]{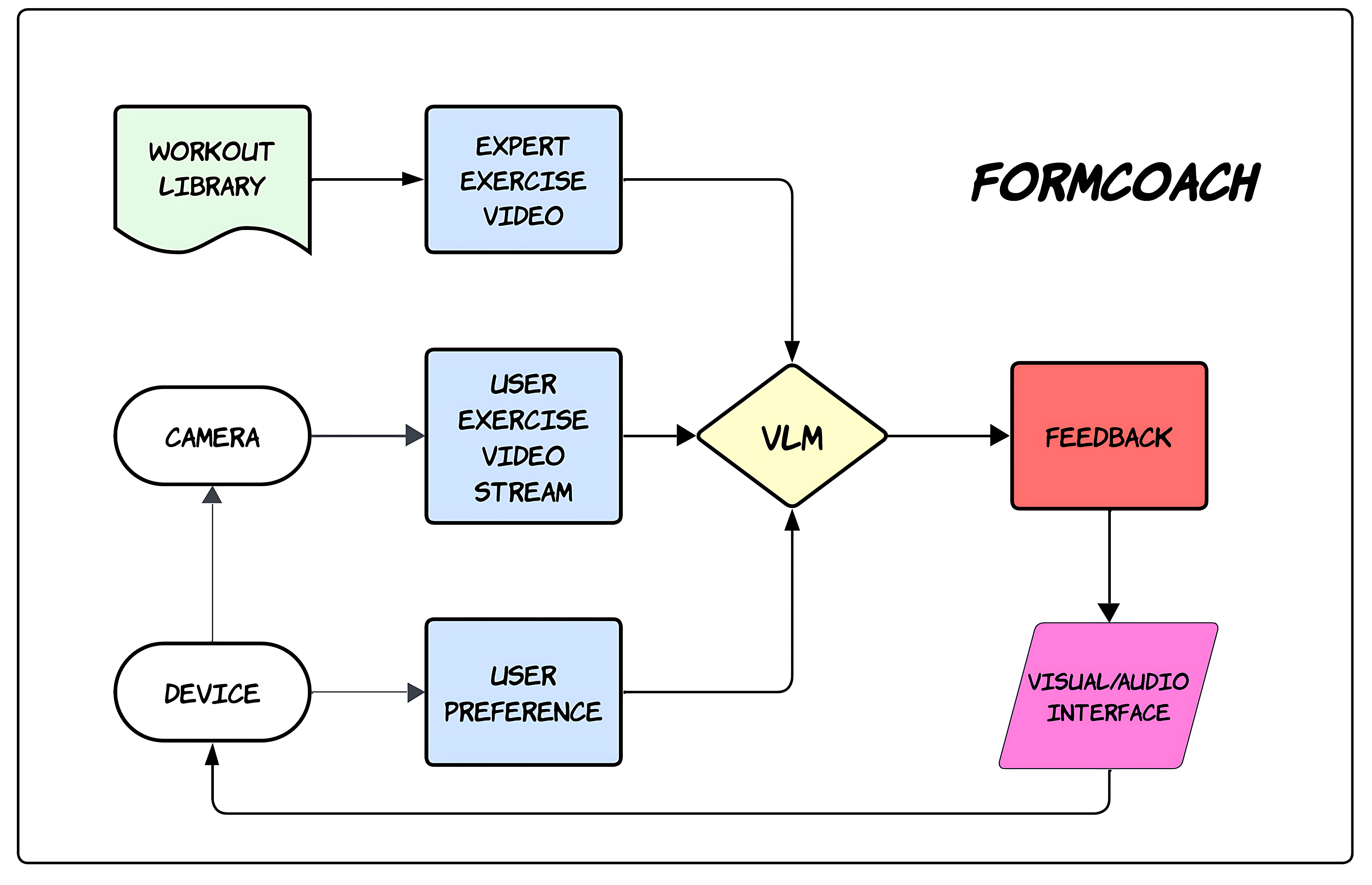}
    \caption{FormCoach pipeline}
    \label{fig:flowchart}
\end{figure}
\section{AI-powered form coaching}
FormCoach transforms a solitary home workout into an interactive coaching session powered by vision–language models that sees, understands, and responds like a human trainer. The FormCoach pipeline is illustrated in Figure~\ref{fig:flowchart}.

\begin{wrapfigure}{r}{0.5\linewidth} 
    \centering
    \includegraphics[width=\linewidth]{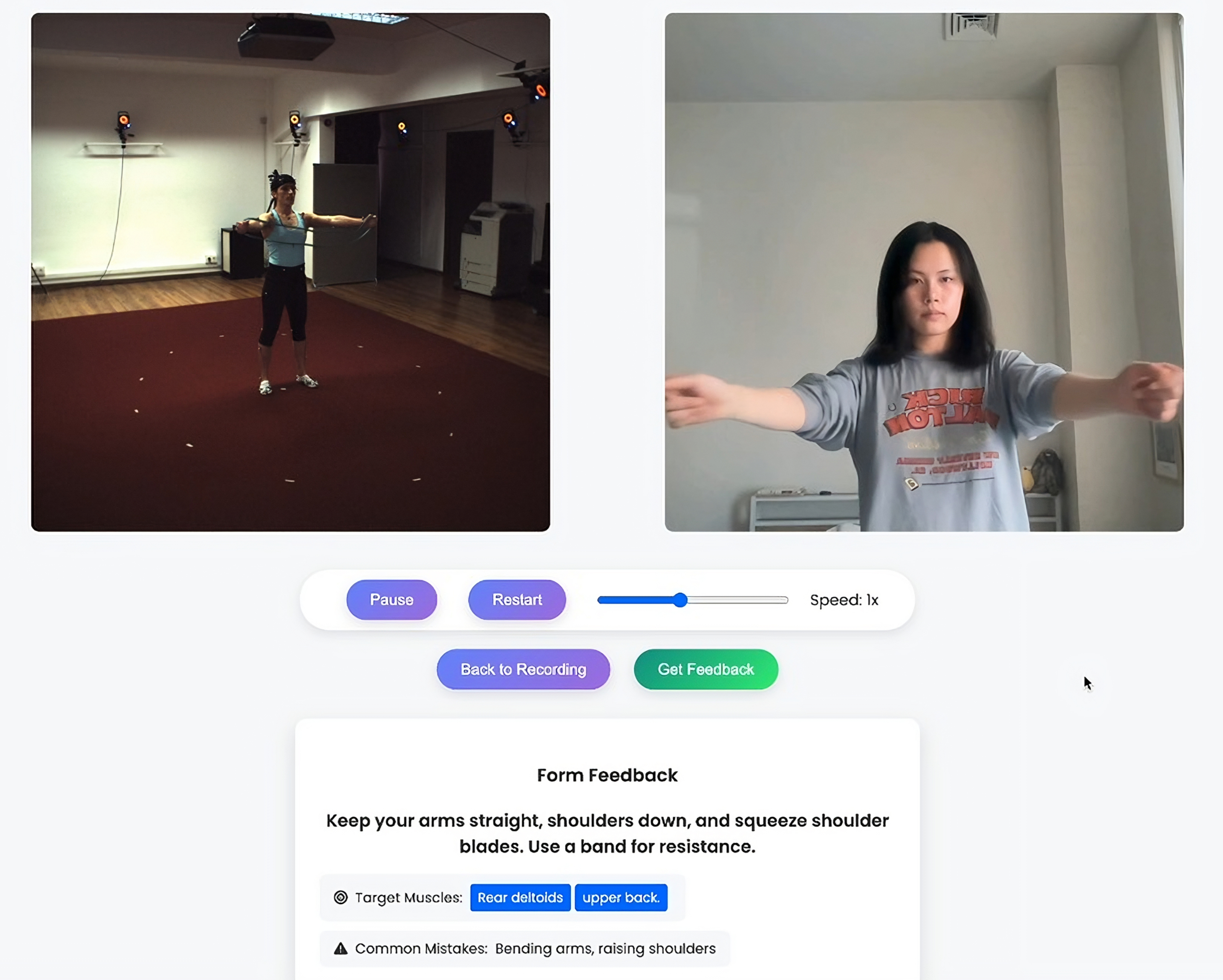}
    \caption{Prototype web interface}
    \label{fig:ui}
\end{wrapfigure}
\textbf{Setup}
The user positions a camera (be it a webcam, smartphone camera, or smart mirror) to capture their workout space. From an extensive, curated library of exercises, the user selects an activity, optionally typing in personal goals or coaching preferences such as “focus on my knee alignment” or “help me improve my depth in squats.” These inputs are incorporated directly into the VLM prompt, shaping the system’s analysis and ensuring feedback is tailored to the user’s priorities. Once configured, FormCoach displays an expert demonstration video alongside the user’s real-time reflection, creating an immediate, side-by-side visual comparison (see Figure~\ref{fig:ui}). 

\textbf{Perform}
As the user begins, FormCoach continuously tracks their body movements, synchronizing frames from the live feed with frames from the expert reference. These frames along with the user's stated preferences, are analyzed by carefully prompted VLMs to detect form deviations.

\textbf{Feedback}
When discrepancies appear, FormCoach responds with targeted, actionable guidance. For instance, if a user's back rounds during a squat, the system might say, "Keep your back straight." Feedback is concise, on-screen, and optionally delivered via text-to-speech for a seamless, hands-free experience that doesn’t disrupt the workout flow.

\section{Benchmark}
\label{sec:baselines}

\subsection{Dataset}

\textbf{Source collection}
Our evaluation dataset uses Fit3D's~\cite{fieraru2021aifit} multi-view exercise recordings. We segmented each sequence into single-rep clips (e.g. one squat). We pair identical exercises to form $10$k reference-user video pairs, each aligned to a fixed-window (mean $\sim$3.8~s) and standardized to $224\times244$ at $30$ FPS. We then uniformly sampled $1,700$ pairs as our evaluation dataset.

\textbf{Exercise coverage}
The dataset covers $22$ exercises, summarized in Table~\ref{tab:exercise_list}, covering upper body lifts (e.g. dumbbell lift overhead), lower body movements (e.g. squats, lunges) and full body exercises (e.g. mule kick, burpee). The dataset is class-balanced, challenging models to perform consistently across exercise types rather than exploiting class frequency. 

\textbf{Ground-truth instructions}
Expert annotators are given the user and reference videos side by side on the gradio interface~\cite{abid2019gradio} and asked to provide concise ($<\!$15 words), actionable corrections comparing the paired movements (e.g., “Keep your back straight throughout the squat.”). These annotations define the FormCoach evaluation set of 1{,}700 pairs and are included in our release.\\

\begin{figure}[H]
  \centering



  \includegraphics[width=0.99\linewidth]{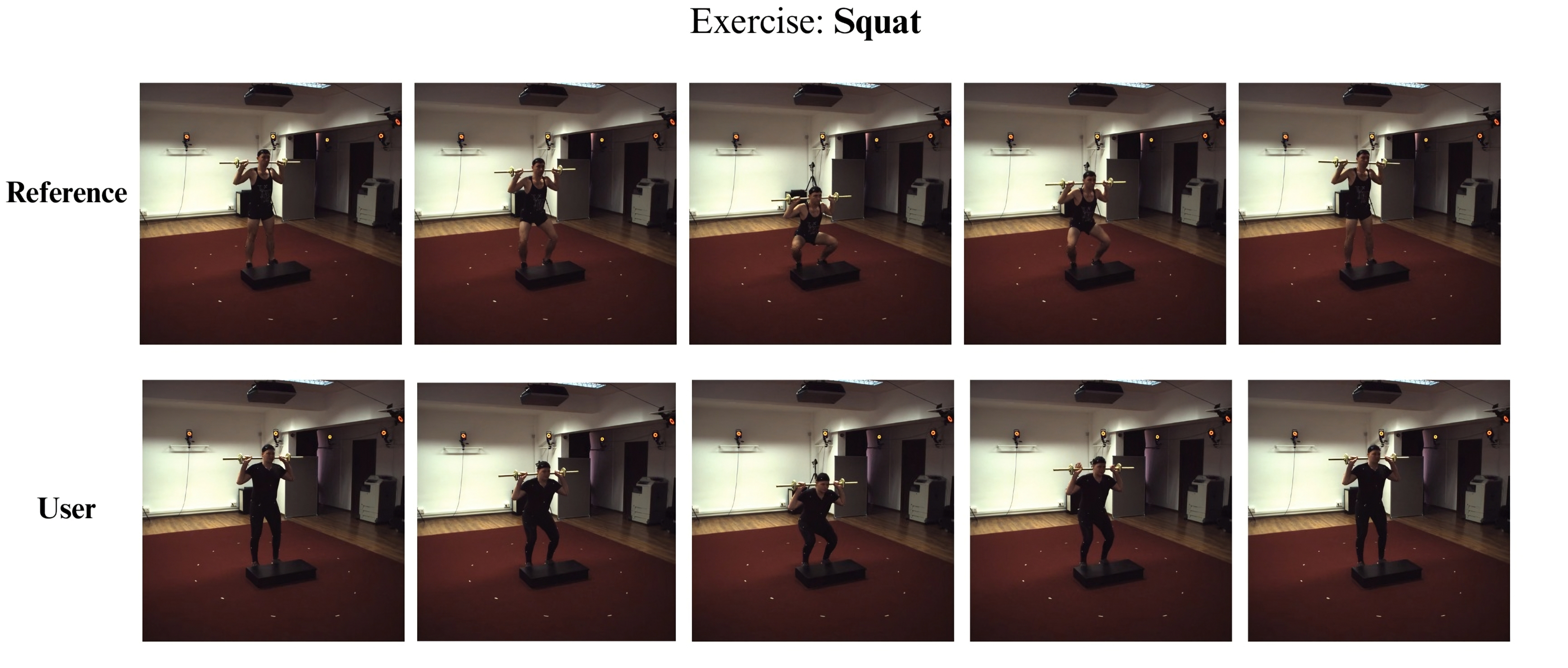}

  \caption{Example from FormCoach evaluation dataset showing the \emph{reference} (top)
           and \emph{user} (bottom) performing the same squatting exercise.  
           Below, we present the expert feedback and sample VLM-generated feedbacks. We highlight the correct feedback in green, hallucinated feedback in yellow, and incorrect/unclear feedback in red.
           }
  \label{fig:example_pair}
  \vspace{0.8em}\hrule\vspace{0.8em}

  \small
  \begin{minipage}{0.9\linewidth}
    \textbf{Expert}\\
    "{\sethlcolor{green}\hl{Spread your feet wider}} and {\sethlcolor{green}\hl{push your knees outward}} as you squat down."
  
    \medskip
    \textbf{GPT‑4.1~\cite{chatgpt2025}} \\
    “{\sethlcolor{green}\hl{Widen your stance}} and {\sethlcolor{green}\hl{push your knees out}} as you squat to improve depth and stability.”

    \medskip
    \textbf{QwenVL2.5-7B~\cite{qwen2}} \\
    "Keep your {\sethlcolor{yellow}\hl{back straight}} and {\sethlcolor{yellow}\hl{core engaged.}}"

    \medskip
    \textbf{VideoLLaMA3~\cite{damonlpsg2025videollama3}} \\
    “Keep your {\sethlcolor{yellow}\hl{chest up}} and {\sethlcolor{yellow}\hl{back straight.}}”
    
    \medskip
    \textbf{Gemini-2.0-~\cite{Hassabis2024Gemini2}} \\
    “Keep your {\sethlcolor{yellow}\hl{chest up}} and ensure your {\sethlcolor{red}\hl{hips break the parallel line}} to maximize glute activation.”

  \end{minipage}
\vspace{0.8em}\hrule\vspace{0.8em}
\end{figure}

\newpage
\subsection{Task Description}

We experiment with various VLMs to generate form feedback for exercise video pairs from our dataset, as seen in Figure~\ref{fig:example_pair}. These models included both API-based models~\cite{chatgpt2025, claude2025, Hassabis2024Gemini2}, and open-source alternatives~\cite{damonlpsg2025videollama3, chen2024internvl, glm2024chatglm, qwen2}. Each query consists of a system prompt and a user-reference exercise video pair. The models are instructed to generate targeted feedback focusing on form corrections within a limit of 15 words. The experimentation encompassed 1,700 exercise pairs, representing a diverse set of exercise types to ensure robust coverage. We ran the API-based models on CPUs and the open-source models on a single NVIDIA A40 GPU with 48 GB memory.

\subsection{Evaluation}
The quality of feedback generated by each model was compared against ground-truth annotations.
We leverage GPT-4.1~\cite{chatgpt2025} for evaluation. GPT is presented with the annotated feedback and the model’s feedback (with relevant context) and asked to rate the model’s feedback on several dimensions. All GPT evaluations are done via the OpenAI API with a deterministic setting (temperature 0) to ensure consistency. The metrics described below:

\begin{table}[h]
\centering
\caption{LLM-based evaluation metrics}
\label{tab:llm_metrics}
\begin{tabular}{p{2.2cm} p{1.2cm} p{7cm}}
\toprule
\textbf{Metric} & {Values} &{Description}\\
\midrule
\textbf{Accuracy} & 5 & Fully correct (pinpoints actual issue)  \\
\textbf{} & 4 & Mostly correct (minor details missing)  \\
\textbf{} & 3 & Partially correct (some missing or incorrect)  \\
\textbf{} & 2 & Mostly incorrect (mostly missing or incorrect)  \\
\textbf{} & 1 & Mostly correct (misses the issue)  \\
\midrule
\textbf{Actionability} & 5 &  Very clear, specific, and actionable\\
\textbf{} & 4 & Mostly clear, specific, and actionable  \\
\textbf{} & 3 & Somewhat clear, specific, and actionable \\
\textbf{} & 2 & Mostly unclear or unsafe  \\
\textbf{} & 1 & Vague or unsafe  \\
\midrule
\textbf{Hallucination} & Yes/No & Whether the model flagged a form error that wasn’t actually present. \\
\bottomrule
\end{tabular}
\end{table}

\section{Results and discussions}
\begin{table}[H]
\centering
\caption{LLM-based evaluation results for different vision-language models.}
\label{tab:llm_eval_results}
\begin{tabular}{lcccc}
\toprule
\textbf{Model} & \textbf{Accuracy (\%)} $\uparrow$ & \textbf{Actionability (\%)} $\uparrow$ & \textbf{Hallucination (\%)} $\downarrow$ \\
\midrule
GPT-4.1~\cite{chatgpt2025}     & \textbf{58.2} & \textbf{94.4} &  \textbf{74.26} \\
GPT-o4-mini~\cite{chatgpt2025o4mini}         & 56.2 & 86.4 &  93.99 \\
Claude Opus 4~\cite{claude2025}   & 49.6 & 84.6 &  82.79 \\
QwenVL2.5-7B~\cite{qwen2}          & 46.6 & 79.8 &  94.39 \\
QwenVL2.5-3B~\cite{qwen2}          & 45.8 & 87.4 &  96.79 \\
InternVL2.5-8B~\cite{chen2024internvl} & 46.4 & 82.4 &  93.60 \\
InternVL2.5-4B~\cite{chen2024internvl} & 44.6 & 87.0 &  95.60 \\
InternVL2.5-2B~\cite{chen2024internvl} & 42.6 & 83.4 &  94.16 \\
GLM4V-9B~\cite{glm2024chatglm}           & 45.0 & 85.6 & 94.62 \\
VideoLLaMA3~\cite{damonlpsg2025videollama3}  & 44.8 & 69.0 &  75.98 \\
Gemini-2.0-Flash~\cite{Hassabis2024Gemini2}         & 43.8 & 91.2 & 97.3 \\

\bottomrule
\end{tabular}
\end{table}



\newpage
\subsection{Performance gaps and opportunities}
While the evaluated models show encouraging potential, they still fall short of expert human feedback. GPT-4.1~\cite{chatgpt2025}, top performer, achieved an accuracy score of 58.2 and an actionability score of 94.4, meaning it consistently delivers clear instructions but correctly identifies only a little over half of actual form issues. The gap highlights room for improvement in precisely identifying and articulating form discrepancies. Other models lagged further behind, with lower accuracy and higher hallucination rates, sometimes inventing corrections unrelated to the reference form or missing mistakes entirely. 

These shortcomings are not surprising given the inherent challenges of video-only analysis. Relying solely on visual input makes the system vulnerable to occlusions (e.g., obscured joints or equipment) and viewpoint limitations, which hinder detection of fine-grained posture errors. Even when movement differences exist, they can appear ambiguous or indistinguishable from certain camera angles.\\ 
These limitations indicate clear directions for enhancement. Multi-modal augmentation, such as combining video with 3D joint position estimation\cite{cao2019openpose, toshev2014deeppose}, could add spatial depth and enhance the model’s sensitivity to subtle kinematic differences. Furthermore, integrating wearable sensors~\cite{zhou2016never}, such as inertial measurement units (IMUs)~\cite{seuter2016live}, can provide precise and viewpoint-independent motion data. These enhancements would substantially reduce ambiguity, boost detection accuracy, and close the gap between AI and expert coaching, unlocking a new level of reliability and trust in AI-powered fitness guidance.

\subsection{Exercise‑wise breakdown}
\begin{figure}[H]
    \centering
    \includegraphics[width=0.8\linewidth]{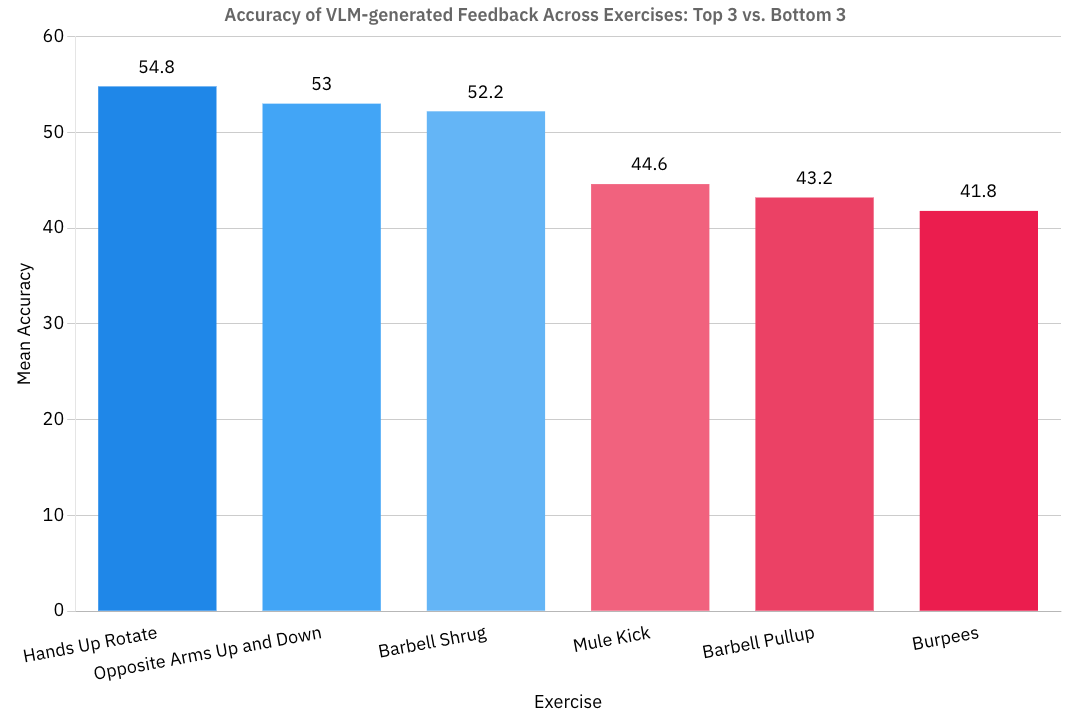}
    \caption{Accuracy of VLM-generated Feedback Across Exercises: Top 3 vs. Bottom 3}
    \label{fig:accuracy}
\end{figure}

We computed the mean accuracy score for each exercise across all evaluated VLMs, highlighting the three highest- and three lowest-performing cases in Figure~\ref{fig:accuracy}. The top performers  (barbell shrug, hands-up rotate, and opposite arms up-and-down warmup) are predominantly isolated, upper-body movements. These exercises are visually straightforward, involving a single limb group with minimal self-occlusion, clear lines of sight, and relatively simple motion patterns. This suggests that VLMs excel when analyzing movements with distinct, easily trackable features and limited inter-limb interference.

In contrast, the lowest-performing exercises (burpees, barbell pull-up, and mule kick) are complex, full-body activities characterized by multiple movement phases, rapid posture transitions, and simultaneous coordination of several limb groups. These dynamics introduce frequent occlusion, abrupt viewpoint changes, and more complex spatial relationships between joints, all of which challenge current VLM architectures. The pronounced drop in accuracy for these tasks underscores an important limitation: today’s VLMs are not yet adept at reliably parsing the fluid, multi-stage mechanics of compound movements.


\section{The future of at-home fitness coaching}
\textbf{From Feedback to Collaboration}: A key opportunity lies in shifting from one-way feedback to two-way collaboration. In its current form, FormCoach primarily tells the user what to fix, but true coaching is often a dialogue. We envision future AI fitness partners that not only give instructions, but also listen to the user's concerns, goals, and progress~\cite{Garbett_Degutyte_Hodge_Astell_2021, Luo_Aguilera_Lyles_Figueroa_2021}. By integrating wearable sensing for real-time state detection with conversational AI, such systems could adapt feedback dynamically, offering not just corrections, but strategic guidance tailored to the user’s intent and physical condition. 

\textbf{Augmented Reality and Embodiment}: Advances in AR hardware and software are rapidly opening the door to immersive, embodied coaching experiences~\cite{arfit2022, pedagogue2025}. In an AR-enhanced FormCoach, lightweight smart glasses could project a virtual instructor directly into the user’s environment, visually overlaying highlights where their form diverges from the reference. Spatial audio could make feedback even more intuitive: imagine hearing your coach’s voice coming from the direction of a misaligned limb. This merging of spatial computing and AI analysis creates the potential for fully embodied, in-the-moment guidance that feels as natural as training with a human partner.

\textbf{Accessible Fitness Coaching}
With AI coaches, expert guidance could become accessible to anyone with a camera, democratizing high-quality fitness training across geographies and income levels~\cite{DemocratizingFitness2025}. FormCoach envisions a future where every home gym includes a built-in AI trainer~\cite{SmartHomeGyms2025}, much like AI assistants are now standard in smart homes~\cite{Brown_2019, Cottier_2024}. The challenge will be to ensure that as these systems become more pervasive, their feedback remains accurate, safe, and personalized—even as users push their physical limits.

\section{Conclusion}
We presented FormCoach, an interactive AI form coach capable of delivering targeted, actionable exercise feedback in real time. Alongside the system, we developed a curated dataset of expert-annotated user–reference exercise pairs and used it to systematically evaluate state-of-the-art vision–language models (VLMs) on nuanced, comparative human motion analysis. Our results demonstrate that VLMs can move beyond generic video understanding toward precise, context-aware guidance, positioning AI as an interactive, creative coaching partner. Our experiments show that while top-performing models such as GPT-4.1 exhibit strong actionability and partial accuracy, they still miss subtle form discrepancies and occasionally hallucinate nonexistent issues. These findings highlight both the promise and the limitations of current VLMs in high-stakes, precision-oriented applications. Overall, FormCoach opens the door to a new class of AI systems that not only see but also engage with users, guiding them through skill acquisition in a collaborative, human-like manner. As multimodal AI advances and incorporates richer sensing, the vision of universally accessible, trustworthy AI coaches moves closer to reality.\\

{
\small
\bibliographystyle{unsrtnat} 
\bibliography{references}     

\begin{thebibliography}{34}
\providecommand{\natexlab}[1]{#1}
\providecommand{\url}[1]{\texttt{#1}}
\expandafter\ifx\csname urlstyle\endcsname\relax
  \providecommand{\doi}[1]{doi: #1}\else
  \providecommand{\doi}{doi: \begingroup \urlstyle{rm}\Url}\fi

\bibitem[Jones et~al.(1994)Jones, Cowan, and Knapik]{Jones_Cowan_Knapik_1994}
Bruce~H. Jones, David~N. Cowan, and Joseph~J. Knapik.
\newblock Exercise, training and injuries*.
\newblock \emph{Sports Medicine}, 18\penalty0 (3):\penalty0 202–214, Sep 1994.
\newblock \doi{10.2165/00007256-199418030-00005}.

\bibitem[Aasa et~al.(2016)Aasa, Svartholm, Andersson, and Berglund]{bjsm2016_sysreview}
Ulrika Aasa, Ivar Svartholm, Fredrik Andersson, and Lars Berglund.
\newblock Injuries among weightlifters and powerlifters: a systematic review.
\newblock \emph{British Journal of Sports Medicine}, 2016.

\bibitem[Tsiouris et~al.(2020)Tsiouris, Tsakanikas, Gatsios, and Fotiadis]{Tsiouris_Tsakanikas_Gatsios_Fotiadis_2020}
Kostas~M. Tsiouris, Vassilios~D. Tsakanikas, Dimitrios Gatsios, and Dimitrios~I. Fotiadis.
\newblock A review of virtual coaching systems in healthcare: Closing the loop with real-time feedback.
\newblock \emph{Frontiers in Digital Health}, 2, Sep 2020.
\newblock \doi{10.3389/fdgth.2020.567502}.

\bibitem[Braga-Pereira et~al.(2024)Braga-Pereira, Furtado, Campos, Sampaio, and Teques]{plosone2024coach_behavior}
R.~Braga-Pereira, G.~E. Furtado, F.~Campos, A.~R. Sampaio, and P.~Teques.
\newblock Impact of fitness coach behavior on exercise motivation, commitment, and enjoyment: A longitudinal study.
\newblock \emph{PLOS ONE}, 2024.

\bibitem[Graves et~al.(2013)Graves, Iyer, Willis, Ebel, Rivara, and Vavilala]{Graves_Iyer_Willis_Ebel_Rivara_Vavilala_2013}
Janessa~M Graves, Krithika~R Iyer, Margaret~M Willis, Beth~E Ebel, Frederick~P Rivara, and Monica~S Vavilala.
\newblock Emergency department-reported injuries associated with mechanical home exercise equipment in the usa.
\newblock \emph{Injury Prevention}, 20\penalty0 (4):\penalty0 281–285, Sep 2013.
\newblock \doi{10.1136/injuryprev-2013-040833}.

\bibitem[Fieraru et~al.(2021)Fieraru, Zanfir, Pirlea, Olaru, and Sminchisescu]{fieraru2021aifit}
Mihai Fieraru, Mihai Zanfir, Silviu-Cristian Pirlea, Vlad Olaru, and Cristian Sminchisescu.
\newblock {AIFit}: Automatic 3d human-interpretable feedback models for fitness training.
\newblock In \emph{IEEE/CVF Conference on Computer Vision and Pattern Recognition (CVPR)}, 2021.

\bibitem[Chen and Yang(2020)]{Chen_Yang_2020}
Steven Chen and Richard~R. Yang.
\newblock Pose trainer: Correcting exercise posture using pose estimation, Jun 2020.
\newblock URL \url{https://arxiv.org/abs/2006.11718}.

\bibitem[Liu et~al.(2024)Liu, Saquib, Zhutian, Kazi, Wei, Fu, and Tai]{Liu_Saquib_Zhutian_Kazi_Wei_Fu_Tai_2024}
Jingyuan Liu, Nazmus Saquib, Chen Zhutian, Rubaiat~Habib Kazi, Li-Yi Wei, Hongbo Fu, and Chiew-Lan Tai.
\newblock Posecoach: A customizable analysis and visualization system for video-based running coaching.
\newblock \emph{IEEE Transactions on Visualization and Computer Graphics}, 30\penalty0 (7):\penalty0 3180–3195, Jul 2024.
\newblock \doi{10.1109/tvcg.2022.3230855}.

\bibitem[Dittakavi et~al.(2022)Dittakavi, Bavikadi, Desai, Chakraborty, Reddy, Balasubramanian, Callepalli, and Sharma]{Dittakavi_Bavikadi_Desai_Chakraborty_Reddy_Balasubramanian_Callepalli_Sharma_2022}
Bhat Dittakavi, Divyagna Bavikadi, Sai~Vikas Desai, Soumi Chakraborty, Nishant Reddy, Vineeth~N Balasubramanian, Bharathi Callepalli, and Ayon Sharma.
\newblock Pose tutor: An explainable system for pose correction in the wild.
\newblock \emph{2022 IEEE/CVF Conference on Computer Vision and Pattern Recognition Workshops (CVPRW)}, page 3539–3548, Jun 2022.
\newblock \doi{10.1109/cvprw56347.2022.00398}.

\bibitem[Athanasiou et~al.(2024)Athanasiou, Ceske, Diomataris, Black, and Varol]{athanasiou2024motionfix}
Nikos Athanasiou, Alp{\'a}r Ceske, Markos Diomataris, Michael~J. Black, and G{\"u}l Varol.
\newblock {MotionFix}: Text-driven 3d human motion editing.
\newblock In \emph{SIGGRAPH Asia 2024 Conference Papers}, 2024.

\bibitem[Zhang et~al.(2024)Zhang, Huang, Liu, Tang, Lu, Chen, Bai, Chu, Yu, and Ouyang]{Zhang_Huang_Liu_Tang_Lu_Chen_Bai_Chu_Yu_Ouyang_2024}
Yaqi Zhang, Di~Huang, Bin Liu, Shixiang Tang, Yan Lu, Lu~Chen, Lei Bai, Qi~Chu, Nenghai Yu, and Wanli Ouyang.
\newblock Motiongpt: Finetuned llms are general-purpose motion generators.
\newblock \emph{Proceedings of the AAAI Conference on Artificial Intelligence}, 38\penalty0 (7):\penalty0 7368–7376, Mar 2024.
\newblock \doi{10.1609/aaai.v38i7.28567}.

\bibitem[Chen et~al.(2024{\natexlab{a}})Chen, Lu, Zeng, Zhang, Wang, Zhang, and Zhang]{motionllm}
Ling-Hao Chen, Shunlin Lu, Ailing Zeng, Hao Zhang, Benyou Wang, Ruimao Zhang, and Lei Zhang.
\newblock Motionllm: Understanding human behaviors from human motions and videos.
\newblock \emph{arxiv:2405.20340}, 2024{\natexlab{a}}.

\bibitem[{OpenAI}(2025{\natexlab{a}})]{chatgpt2025}
{OpenAI}.
\newblock Chatgpt (gpt-4), 2025{\natexlab{a}}.
\newblock URL \url{https://chat.openai.com}.
\newblock Large language model.

\bibitem[Boqiang~Zhang(2025)]{damonlpsg2025videollama3}
Zesen Cheng Zhiqiang Hu Yuqian Yuan Guanzheng Chen Sicong Leng Yuming Jiang Hang Zhang Xin Li Peng Jin Wenqi Zhang Fan Wang Lidong Bing Deli~Zhao Boqiang~Zhang, Kehan~Li.
\newblock Videollama 3: Frontier multimodal foundation models for image and video understanding.
\newblock \emph{arXiv preprint arXiv:2501.13106}, 2025.
\newblock URL \url{https://arxiv.org/abs/2501.13106}.

\bibitem[Chen et~al.(2024{\natexlab{b}})Chen, Wu, Wang, Su, Chen, Xing, Zhong, Zhang, Zhu, Lu, et~al.]{chen2024internvl}
Zhe Chen, Jiannan Wu, Wenhai Wang, Weijie Su, Guo Chen, Sen Xing, Muyan Zhong, Qinglong Zhang, Xizhou Zhu, Lewei Lu, et~al.
\newblock Internvl: Scaling up vision foundation models and aligning for generic visual-linguistic tasks.
\newblock In \emph{Proceedings of the IEEE/CVF Conference on Computer Vision and Pattern Recognition}, pages 24185--24198, 2024{\natexlab{b}}.

\bibitem[Yang et~al.(2024)Yang, Yang, Hui, Zheng, Yu, Zhou, Li, Li, Liu, Huang, Dong, Wei, Lin, Tang, Wang, Yang, Tu, Zhang, Ma, Xu, Zhou, Bai, He, Lin, Dang, Lu, Chen, Yang, Li, Xue, Ni, Zhang, Wang, Peng, Men, Gao, Lin, Wang, Bai, Tan, Zhu, Li, Liu, Ge, Deng, Zhou, Ren, Zhang, Wei, Ren, Fan, Yao, Zhang, Wan, Chu, Liu, Cui, Zhang, and Fan]{qwen2}
An~Yang, Baosong Yang, Binyuan Hui, Bo~Zheng, Bowen Yu, Chang Zhou, Chengpeng Li, Chengyuan Li, Dayiheng Liu, Fei Huang, Guanting Dong, Haoran Wei, Huan Lin, Jialong Tang, Jialin Wang, Jian Yang, Jianhong Tu, Jianwei Zhang, Jianxin Ma, Jin Xu, Jingren Zhou, Jinze Bai, Jinzheng He, Junyang Lin, Kai Dang, Keming Lu, Keqin Chen, Kexin Yang, Mei Li, Mingfeng Xue, Na~Ni, Pei Zhang, Peng Wang, Ru~Peng, Rui Men, Ruize Gao, Runji Lin, Shijie Wang, Shuai Bai, Sinan Tan, Tianhang Zhu, Tianhao Li, Tianyu Liu, Wenbin Ge, Xiaodong Deng, Xiaohuan Zhou, Xingzhang Ren, Xinyu Zhang, Xipin Wei, Xuancheng Ren, Yang Fan, Yang Yao, Yichang Zhang, Yu~Wan, Yunfei Chu, Yuqiong Liu, Zeyu Cui, Zhenru Zhang, and Zhihao Fan.
\newblock Qwen2 technical report.
\newblock \emph{arXiv preprint arXiv:2407.10671}, 2024.

\bibitem[Cho et~al.(2025)Cho, Lin, Srinivasan, Saxon, Kwon, Chavez, and May]{Cho_Lin_Srinivasan_Saxon_Kwon_Chavez_May_2025}
Hyundong~Justin Cho, Spencer Lin, Tejas Srinivasan, Michael Saxon, Deuksin Kwon, Natali~T. Chavez, and Jonathan May.
\newblock Can vision language models understand mimed actions?
\newblock \emph{Findings of the Association for Computational Linguistics: ACL 2025}, page 26744–26759, 2025.
\newblock \doi{10.18653/v1/2025.findings-acl.1372}.

\bibitem[Abid et~al.(2019)Abid, Abdalla, Abid, Khan, Alfozan, and Zou]{abid2019gradio}
Abubakar Abid, Ali Abdalla, Ali Abid, Dawood Khan, Abdulrahman Alfozan, and James Zou.
\newblock Gradio: Hassle-free sharing and testing of ml models in the wild.
\newblock \emph{arXiv preprint arXiv:1906.02569}, 2019.

\bibitem[Hassabis et~al.(2024)Hassabis, Kavukcuoglu, and the DeepMind Gemini~Team]{Hassabis2024Gemini2}
Demis Hassabis, Koray Kavukcuoglu, and the DeepMind Gemini~Team.
\newblock Introducing gemini 2.0: our new ai model for the agentic era.
\newblock Google DeepMind blog, 2024.
\newblock URL \url{https://blog.google/technology/google-deepmind/google-gemini-ai-update-december-2024}.
\newblock December 11, 2024.

\bibitem[{Anthropic}(2025)]{claude2025}
{Anthropic}.
\newblock Claude opus 4, 2025.
\newblock URL \url{https://claude.ai}.
\newblock Large language model.

\bibitem[GLM et~al.(2024)GLM, Zeng, Xu, Wang, Zhang, Yin, Rojas, Feng, Zhao, Lai, Yu, Wang, Sun, Zhang, Cheng, Gui, Tang, Zhang, Li, Zhao, Wu, Zhong, Liu, Huang, Zhang, Zheng, Lu, Duan, Zhang, Cao, Yang, Tam, Zhao, Liu, Xia, Zhang, Gu, Lv, Liu, Liu, Yang, Song, Zhang, An, Xu, Niu, Yang, Li, Bai, Dong, Qi, Wang, Yang, Du, Hou, and Wang]{glm2024chatglm}
Team GLM, Aohan Zeng, Bin Xu, Bowen Wang, Chenhui Zhang, Da~Yin, Diego Rojas, Guanyu Feng, Hanlin Zhao, Hanyu Lai, Hao Yu, Hongning Wang, Jiadai Sun, Jiajie Zhang, Jiale Cheng, Jiayi Gui, Jie Tang, Jing Zhang, Juanzi Li, Lei Zhao, Lindong Wu, Lucen Zhong, Mingdao Liu, Minlie Huang, Peng Zhang, Qinkai Zheng, Rui Lu, Shuaiqi Duan, Shudan Zhang, Shulin Cao, Shuxun Yang, Weng~Lam Tam, Wenyi Zhao, Xiao Liu, Xiao Xia, Xiaohan Zhang, Xiaotao Gu, Xin Lv, Xinghan Liu, Xinyi Liu, Xinyue Yang, Xixuan Song, Xunkai Zhang, Yifan An, Yifan Xu, Yilin Niu, Yuantao Yang, Yueyan Li, Yushi Bai, Yuxiao Dong, Zehan Qi, Zhaoyu Wang, Zhen Yang, Zhengxiao Du, Zhenyu Hou, and Zihan Wang.
\newblock Chatglm: A family of large language models from glm-130b to glm-4 all tools.
\newblock \emph{arXiv}, 2024.

\bibitem[{OpenAI}(2025{\natexlab{b}})]{chatgpt2025o4mini}
{OpenAI}.
\newblock Chatgpt (gpt-4), 2025{\natexlab{b}}.
\newblock URL \url{https://chat.openai.com}.
\newblock Large language model.

\bibitem[Cao et~al.(2019)Cao, Hidalgo, Simon, Wei, and Sheikh]{cao2019openpose}
Zhe Cao, Gines Hidalgo, Tomas Simon, Shih-En Wei, and Yaser Sheikh.
\newblock Openpose: Realtime multi-person 2d pose estimation using part affinity fields.
\newblock \emph{IEEE transactions on pattern analysis and machine intelligence}, 43\penalty0 (1):\penalty0 172--186, 2019.

\bibitem[Toshev and Szegedy(2014)]{toshev2014deeppose}
Alexander Toshev and Christian Szegedy.
\newblock Deeppose: Human pose estimation via deep neural networks.
\newblock In \emph{Proceedings of the IEEE conference on computer vision and pattern recognition}, pages 1653--1660, 2014.

\bibitem[Zhou et~al.(2016)Zhou, Sundholm, Cheng, Cruz, and Lukowicz]{zhou2016never}
Bo~Zhou, Mathias Sundholm, Jingyuan Cheng, Heber Cruz, and Paul Lukowicz.
\newblock Never skip leg day: A novel wearable approach to monitoring gym leg exercises.
\newblock In \emph{2016 IEEE International Conference on Pervasive Computing and Communications (PerCom)}, pages 1--9. IEEE, 2016.

\bibitem[Seuter et~al.(2016)Seuter, Opitz, Bauer, and Hochmann]{seuter2016live}
Matthias Seuter, Lucien Opitz, Gernot Bauer, and David Hochmann.
\newblock Live-feedback from the imus: Animated 3d visualization for everyday-exercising.
\newblock In \emph{Proceedings of the 2016 ACM International Joint Conference on Pervasive and Ubiquitous Computing: Adjunct}, pages 904--907, 2016.

\bibitem[Garbett et~al.(2021)Garbett, Degutyte, Hodge, and Astell]{Garbett_Degutyte_Hodge_Astell_2021}
Andrew Garbett, Ziedune Degutyte, James Hodge, and Arlene Astell.
\newblock Towards understanding people’s experiences of ai computer vision fitness instructor apps.
\newblock \emph{Designing Interactive Systems Conference 2021}, page 1619–1637, Jun 2021.
\newblock \doi{10.1145/3461778.3462094}.

\bibitem[Luo et~al.(2021)Luo, Aguilera, Lyles, and Figueroa]{Luo_Aguilera_Lyles_Figueroa_2021}
Tiffany~Christina Luo, Adrian Aguilera, Courtney~Rees Lyles, and Caroline~Astrid Figueroa.
\newblock Promoting physical activity through conversational agents: Mixed methods systematic review.
\newblock \emph{Journal of Medical Internet Research}, 23\penalty0 (9), Sep 2021.
\newblock \doi{10.2196/25486}.

\bibitem[Ghosh et~al.(2022)Ghosh, Nguyen, Khan, and Al~Moubayed]{arfit2022}
Suparna Ghosh, Thanh Nguyen, Wasif Khan, and Noura Al~Moubayed.
\newblock Arfit: An augmented reality-based virtual trainer for physical exercise learning.
\newblock \emph{arXiv preprint arXiv:2209.02161}, 2022.
\newblock URL \url{https://arxiv.org/abs/2209.02161}.

\bibitem[{Pedagogue}(2025)]{pedagogue2025}
{Pedagogue}.
\newblock Augmented reality for real-time coaching feedback.
\newblock \url{https://pedagogue.app/augmented-reality-for-real-time-coaching-feedback/}, 2025.
\newblock Accessed: 2025-08-09.

\bibitem[Magazine(2025)]{DemocratizingFitness2025}
AllTech Magazine.
\newblock Bringing the personal trainer home: How ai is democratizing fitness.
\newblock \url{https://alltechmagazine.com/how-ai-is-democratizing-fitness/}, 2025.
\newblock Accessed: 2025-08-09.

\bibitem[Engineer(2025)]{SmartHomeGyms2025}
Fitness Engineer.
\newblock Ai smart home gyms 2025: Future of intelligent fitness technology.
\newblock \url{https://ai-fitness-engineer.com/the-future-of-ai-in-smart-home-gyms}, 2025.
\newblock Accessed: 2025-08-09.

\bibitem[Brown(2019)]{Brown_2019}
Bruce Brown.
\newblock Alexa and google home smart speakers bring a.i. to nearly one in three u.s. homes, May 2019.
\newblock URL \url{https://www.digitaltrends.com/home/alexa-and-google-home-smart-speakers-bring-ai-to-one-in-three-us-homes/}.

\bibitem[Cottier(2024)]{Cottier_2024}
Emilie Cottier.
\newblock Smart home assistants: How iot is transforming everyday living, Dec 2024.
\newblock URL \url{https://thethings.io/iot-solutions/smart-home-assistants-how-iot-is-transforming-everyday-living/}.

\end{thebibliography}
}
\medskip

\newpage
\appendix

\section{Technical Appendices and Supplementary Material}

\subsection{Dataset}
The dataset includes 22 common strength and mobility exercises as listed in Table~\ref{tab:exercise_list}. The dataset is hosted on Huggingface and can be accessed \href{https://huggingface.co/datasets/youdahua/fit3d}{Here}.

\begin{table}[H]
\centering
\caption{List of Exercises Included in the Dataset}
\label{tab:exercise_list}
\begin{tabular}{cl}
\toprule
\textbf{No.} & \textbf{Exercise Name} \\
\midrule
1 & Dumbbell Lift Overhead             \\
2 & Dumbbell Lunge Backward            \\
3 & Hands Up Rotate       \\
4 & Barbell Pull Up        \\
5 & Clean and Press           \\
6 & Dumbbell Lateral Raise             \\
7 & Burpee    \\
8 & Up and Down Stretch          \\
9 & Dumbbell Chest Fly          \\
10 & Arms Raise Overhead         \\
11 & Standing Ab Twists        \\
12 & Opposite Arms Up and Down Warmup        \\
13 & Dumbbell Squat and Press \\
14 & Lunge and Reach Hands Front     \\
15 & Mule Kick  \\ 
16 & Arms Circles\\
17 & Squat    \\
18 & One Arm Row       \\
19 & Squat and Jump with Hands Up          \\
20 & Dumbbell High Pulls\\
21 & Deadlift    \\
22 & Barbell Shrug \\
\bottomrule
\end{tabular}
\end{table}
 \newpage
\subsection{Ground Truth Instruction Annotations}
\begin{figure}[H]
    \centering
    \includegraphics[width=0.8\linewidth]{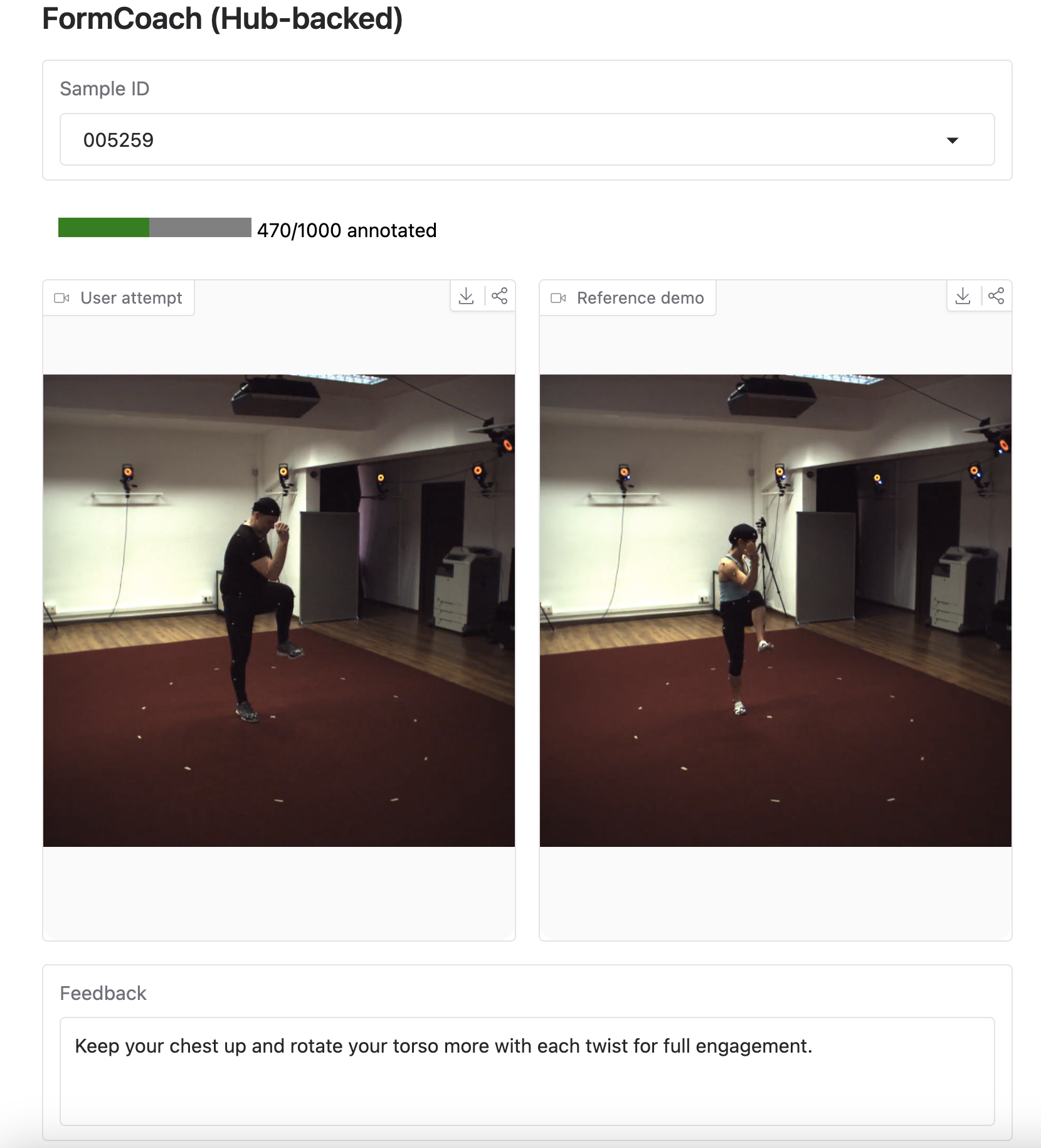}
    \caption{Instruction Annotation Interface on Gradio}
    \label{fig:gradio}
\end{figure}

Experts annotated ground truth instructions given the user and reference exercise pairs through the Gradio interface as illustrated in Figure ~\ref{fig:gradio}.

\subsection{Prompts}

\begin{figure}[H]
    \centering
    \includegraphics[width=0.95\linewidth]{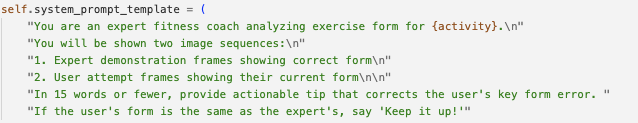}
    \caption{Prompt for feedback generation}
    \label{fig:prompt_gen}
\end{figure}

\begin{figure}[H]
    \centering
    \includegraphics[width=0.95\linewidth]{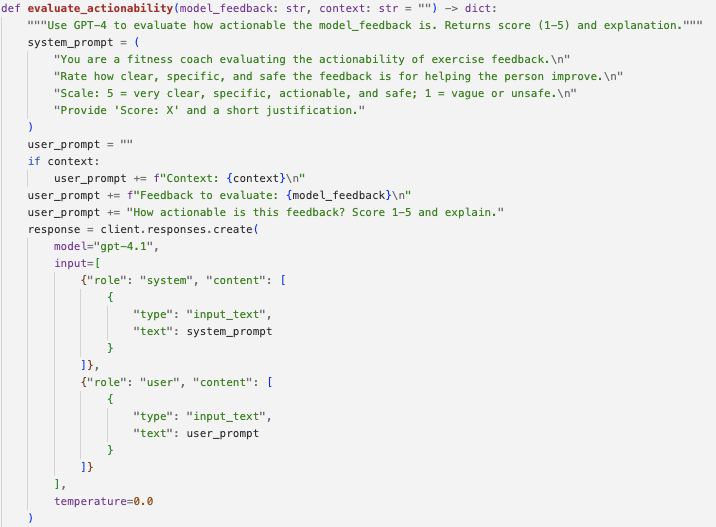}
    \caption{Prompt for actionability evaluation}
    \label{fig:act_eval}
\end{figure}

\begin{figure}[H]
    \centering
    \includegraphics[width=0.95\linewidth]{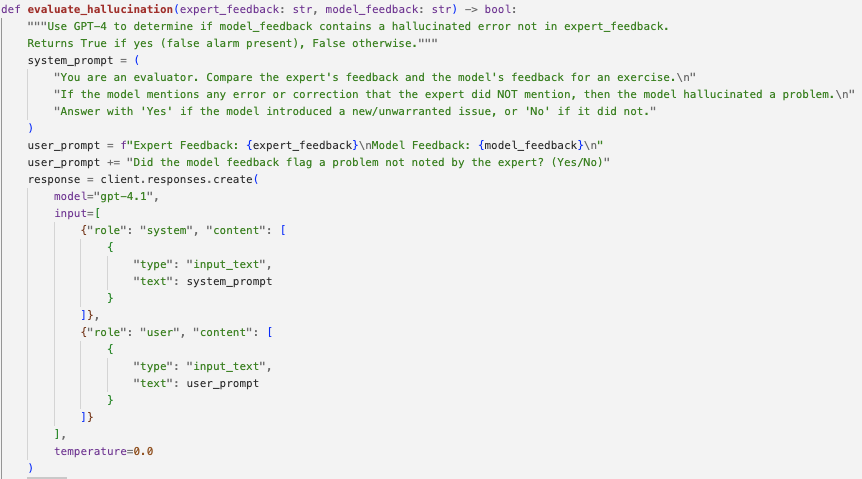}
    \caption{Prompt for hallucination evaluation}
    \label{fig:hal_eval}
\end{figure}

Prompts for feedback generation and evaluations are shown in Figure~\ref{fig:prompt_gen}, Figure~\ref{fig:act_eval}, Figure~\ref{fig:act_eval}, and Figure~\ref{fig:hal_eval}.
\begin{figure}[H]
    \centering
    \includegraphics[width=0.95\linewidth]{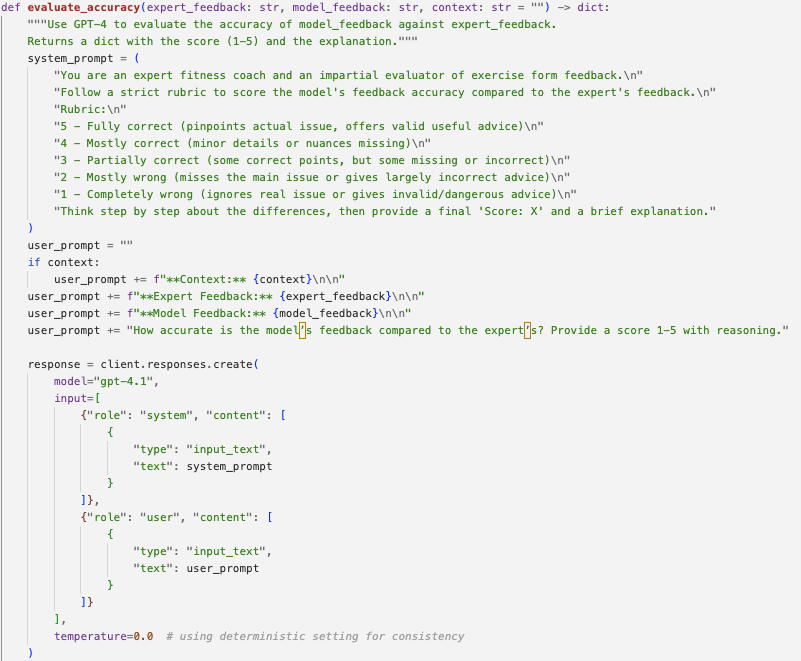}
    \caption{Prompt for accuracy evaluation}
    \label{fig:acc_eval}
\end{figure}


\newpage
\section*{NeurIPS Paper Checklist}

The checklist is designed to encourage best practices for responsible machine learning research, addressing issues of reproducibility, transparency, research ethics, and societal impact. Do not remove the checklist: {\bf The papers not including the checklist will be desk rejected.} The checklist should follow the references and follow the (optional) supplemental material.  The checklist does NOT count towards the page
limit. 

Please read the checklist guidelines carefully for information on how to answer these questions. For each question in the checklist:
\begin{itemize}
    \item You should answer \answerYes{}, \answerNo{}, or \answerNA{}.
    \item \answerNA{} means either that the question is Not Applicable for that particular paper or the relevant information is Not Available.
    \item Please provide a short (1–2 sentence) justification right after your answer (even for NA). 
\end{itemize}

{\bf The checklist answers are an integral part of your paper submission.} They are visible to the reviewers, area chairs, senior area chairs, and ethics reviewers. You will be asked to also include it (after eventual revisions) with the final version of your paper, and its final version will be published with the paper.

The reviewers of your paper will be asked to use the checklist as one of the factors in their evaluation. While "\answerYes{}" is generally preferable to "\answerNo{}", it is perfectly acceptable to answer "\answerNo{}" provided a proper justification is given (e.g., "error bars are not reported because it would be too computationally expensive" or "we were unable to find the license for the dataset we used"). In general, answering "\answerNo{}" or "\answerNA{}" is not grounds for rejection. While the questions are phrased in a binary way, we acknowledge that the true answer is often more nuanced, so please just use your best judgment and write a justification to elaborate. All supporting evidence can appear either in the main paper or the supplemental material, provided in appendix. If you answer \answerYes{} to a question, in the justification please point to the section(s) where related material for the question can be found.

IMPORTANT, please:
\begin{itemize}
    \item {\bf Delete this instruction block, but keep the section heading ``NeurIPS Paper Checklist"},
    \item  {\bf Keep the checklist subsection headings, questions/answers and guidelines below.}
    \item {\bf Do not modify the questions and only use the provided macros for your answers}.
\end{itemize}


\begin{enumerate}

\item {\bf Claims}
    \item[] Question: Do the main claims made in the abstract and introduction accurately reflect the paper's contributions and scope?
    \item[] Answer: \answerYes{} 
    \item[] Justification: {The abstract and introduction describes the motivation, background, and main contributions of FormCoach which is an AI fitness coach powered by VLMs.}

\item {\bf Limitations}
    \item[] Question: Does the paper discuss the limitations of the work performed by the authors?
    \item[] Answer: \answerYes{} 
    \item[] Justification: {Section 4.1 and Section 5 discusses the limitation of video-only approach and future directions of improving form feedback accuracy.}

\item {\bf Theory assumptions and proofs}
    \item[] Question: For each theoretical result, does the paper provide the full set of assumptions and a complete (and correct) proof?
    \item[] Answer: \answerYes{} 
    \item[] Justification: {FormCoach provides the prompts for generating feedback and details about how the test dataset is generated and feedbacks are evaluated.}

    \item {\bf Experimental result reproducibility}
    \item[] Question: Does the paper fully disclose all the information needed to reproduce the main experimental results of the paper to the extent that it affects the main claims and/or conclusions of the paper (regardless of whether the code and data are provided or not)?
    \item[] Answer: \answerYes{} 
    \item[] Justification: {The gradio demo space is public for viewing the dataset. We'll release a repo for the evaluation pipeline and the web demo soon.}

\item {\bf Open access to data and code}
    \item[] Question: Does the paper provide open access to the data and code, with sufficient instructions to faithfully reproduce the main experimental results, as described in supplemental material?
    \item[] Answer: \answerYes{} 
    \item[] Justification: {The data can be accessed on the gradio app and the code can be accessed on github as mentioned in appendix.}

\item {\bf Experimental setting/details}
    \item[] Question: Does the paper specify all the training and test details (e.g., data splits, hyperparameters, how they were chosen, type of optimizer, etc.) necessary to understand the results?
    \item[] Answer: \answerYes{} 
    \item[] Justification: {The evaluation data, models and prompts are provided.}

\item {\bf Experiment statistical significance}
    \item[] Question: Does the paper report error bars suitably and correctly defined or other appropriate information about the statistical significance of the experiments?
    \item[] Answer: \answerYes{} 
    \item[] Justification: {Evaluation metrics are thoroughly explained and results are discussed. }

\item {\bf Experiments compute resources}
    \item[] Question: For each experiment, does the paper provide sufficient information on the computer resources (type of compute workers, memory, time of execution) needed to reproduce the experiments?
    \item[] Answer: \answerYes{} 
    \item[] Justification: {Section 4.2 mentions the resources used.}
    
\item {\bf Code of ethics}
    \item[] Question: Does the research conducted in the paper conform, in every respect, with the NeurIPS Code of Ethics \url{https://neurips.cc/public/EthicsGuidelines}?
    \item[] Answer: \answerYes{Yes} 
    \item[] Justification: {Anonymity is preserved.}

\item {\bf Broader impacts}
    \item[] Question: Does the paper discuss both potential positive societal impacts and negative societal impacts of the work performed?
    \item[] Answer: \answerYes{} 
    \item[] Justification: {Broader impacts are discussed.}
    
\item {\bf Safeguards}
    \item[] Question: Does the paper describe safeguards that have been put in place for responsible release of data or models that have a high risk for misuse (e.g., pretrained language models, image generators, or scraped datasets)?
    \item[] Answer: \answerNA{} 
    \item[] Justification: {This paper uses publicly open datasets and models and have no such concerns.}

\item {\bf Licenses for existing assets}
    \item[] Question: Are the creators or original owners of assets (e.g., code, data, models), used in the paper, properly credited and are the license and terms of use explicitly mentioned and properly respected?
    \item[] Answer: \answerYes{}{} 
    \item[] Justification: {The assets used are properly cited.}

\item {\bf New assets}
    \item[] Question: Are new assets introduced in the paper well documented and is the documentation provided alongside the assets?
    \item[] Answer: \answerYes{} 
    \item[] Justification: {The dataset is hosted on huggingface.}

\item {\bf Crowdsourcing and research with human subjects}
    \item[] Question: For crowdsourcing experiments and research with human subjects, does the paper include the full text of instructions given to participants and screenshots, if applicable, as well as details about compensation (if any)? 
    \item[] Answer: \answerYes{} 
    \item[] Justification: {The annotator UI is provided in the appendix.}

\item {\bf Institutional review board (IRB) approvals or equivalent for research with human subjects}
    \item[] Question: Does the paper describe potential risks incurred by study participants, whether such risks were disclosed to the subjects, and whether Institutional Review Board (IRB) approvals (or an equivalent approval/review based on the requirements of your country or institution) were obtained?
    \item[] Answer: \answerNA{}{} 
    \item[] Justification: {IRB not required for our study on publicly available dataset.}

\item {\bf Declaration of LLM usage}
    \item[] Question: Does the paper describe the usage of LLMs if it is an important, original, or non-standard component of the core methods in this research? Note that if the LLM is used only for writing, editing, or formatting purposes and does not impact the core methodology, scientific rigorousness, or originality of the research, declaration is not required.
    \item[] Answer: \answerYes{} 
    \item[] Justification: {Usage of LLM for evaluation is explained in the paper.}

\end{enumerate}

\end{document}